\documentclass[sigconf]{acmart}

\AtBeginDocument{%
  \providecommand\BibTeX{{%
    \normalfont B\kern-0.5em{\scshape i\kern-0.25em b}\kern-0.8em\TeX}}}

\usepackage{xspace}
\usepackage{multicol}
\usepackage{multirow}
\usepackage{caption}
\usepackage{subcaption}
\usepackage[group-separator={,},group-minimum-digits={3}]{siunitx}

\usepackage{fancyhdr}

\newcommand{\ours}{\textsc{FS-G}\xspace}
\newcommand{\ourcore}{\textsc{FederatedScope}\xspace}
\newcommand{\subjabbr}{\textsc{FGL}\xspace}

\copyrightyear{2022}
\acmYear{2022}
\setcopyright{acmcopyright}\acmConference[KDD '22]{Proceedings of the 28th ACM SIGKDD Conference on Knowledge Discovery and Data Mining}{August 14--18, 2022}{Washington, DC, USA}
\acmBooktitle{Proceedings of the 28th ACM SIGKDD Conference on Knowledge Discovery and Data Mining (KDD '22), August 14--18, 2022, Washington, DC, USA}
\acmPrice{15.00}
\acmDOI{10.1145/3534678.3539112}
\acmISBN{978-1-4503-9385-0/22/08}

\begin{document}

\title{FederatedScope-GNN: Towards a Unified, Comprehensive and Efficient Package for Federated Graph Learning}

\author{Zhen Wang}
 \affiliation{
   \institution{Alibaba Group}
    \country{}}
 \email{jones.wz@alibaba-inc.com}
 
\author{Weirui Kuang}
 \affiliation{
   \institution{Alibaba Group}
    \country{}}
 \email{weirui.kwr@alibaba-inc.com}

\author{Yuexiang Xie}
\affiliation{
     \institution{Alibaba Group}
      \country{}}
 \email{yuexiang.xyx@alibaba-inc.com}

\author{Liuyi Yao}
\affiliation{%
\institution{Alibaba Group}
    \country{}}
\email{yly287738@alibaba-inc.com}

 \author{Yaliang Li$^*$}
 \affiliation{
   \institution{Alibaba Group}
   \country{}}
 \email{yaliang.li@alibaba-inc.com}

 \author{Bolin Ding}
 \affiliation{
   \institution{Alibaba Group}
   \country{}}
 \email{bolin.ding@alibaba-inc.com}
 
 \author{Jingren Zhou}
 \affiliation{
   \institution{Alibaba Group}
   \country{}}
 \email{jingren.zhou@alibaba-inc.com}
 
\renewcommand{\authors}{Zhen Wang, Weirui Kuang, Yuexiang Xie, Liuyi Yao, Yaliang Li, Bolin Ding, Jingren Zhou}
\renewcommand{\shortauthors}{Zhen Wang et al.}

\begin{abstract}
The incredible development of federated learning (FL) has benefited various tasks in the domains of computer vision and natural language processing, and the existing frameworks such as TFF and FATE has made the deployment easy in real-world applications.
However, federated graph learning (\subjabbr), even though graph data are prevalent, has not been well supported due to its unique characteristics and requirements. The lack of \subjabbr-related framework increases the efforts for accomplishing reproducible research and deploying in real-world applications.
Motivated by such strong demand, in this paper, we first discuss the challenges in creating an easy-to-use \subjabbr package and accordingly present our implemented package \textbf{F}ederated\textbf{S}cope-\textbf{G}NN (\ours), which provides (1) a unified view for modularizing and expressing \subjabbr algorithms; (2) comprehensive DataZoo and ModelZoo for out-of-the-box \subjabbr capability; (3) an efficient model auto-tuning component; and (4) off-the-shelf privacy attack and defense abilities. 
We validate the effectiveness of \ours by conducting extensive experiments, which simultaneously gains many valuable insights about \subjabbr for the community. Moreover, we employ \ours to serve the \subjabbr application in real-world E-commerce scenarios, where the attained improvements indicate great potential business benefits. We publicly release \ours, as submodules of FederatedScope, at \href{https://github.com/alibaba/FederatedScope}{https://github.com/alibaba/FederatedScope} to promote \subjabbr's research and enable broad applications that would otherwise be infeasible due to the lack of a dedicated package.
\end{abstract}



\maketitle

\renewcommand*{\thefootnote}{\fnsymbol{footnote}}
\footnotetext[1]{Corresponding author.}
\renewcommand*{\thefootnote}{\arabic{footnote}}

\section{Introduction}
\label{sec:intro}

Along with the rising concerns about privacy, federated learning (FL)~\cite{def_fl}, a paradigm for collaboratively learning models without access to dispersed data, has attracted more and more attention from both industry and academia.
Its successful applications include keyboard prediction~\cite{keyboard}, object detection~\cite{objectdetection}, speech recognition~\cite{fedspeech}, the list goes on.
This fantastic progress benefits from the FL frameworks, e.g., TFF~\cite{tff} and FATE~\cite{fate}, which save practitioners from the implementation details and facilitate the transfer from research prototype to deployed service.

However, such helpful supports have mainly focused on tasks in vision and language domains. Yet, the graph data, ubiquitous in real-world applications, e.g., recommender systems~\cite{fedgnn}, healthcare~\cite{fedsage}, and anti-money laundering~\cite{AML}, have not been well supported.
As a piece of evidence, most existing FL frameworks, including TFF, FATE, and PySyft~\cite{pysyft}, have not provided off-the-shelf federated graph learning (\subjabbr) capacities, not to mention the lack of FGL benchmarks on a par with LEAF~\cite{leaf} for vision and language tasks.

As a result, FL optimization algorithms, including FedAvg~\cite{def_fl}, FedProx~\cite{fedprox}, and FedOPT~\cite{fedopt}, are mainly evaluated on vision and language tasks.
When applied to optimize graph neural network (GNN) models, their characteristics are unclear to the community.
Another consequence of the lack of dedicated framework support is that many recent \subjabbr works (e.g., FedSage+~\cite{fedsage} and GCFL~\cite{gcfl}) have to implement their methods from scratch and conduct experiments on respective testbeds.

We notice that such a lack of widely-adopted benchmarks and unified implementations of related works have become obstacles to developing novel \subjabbr methods and the deployment in real-world applications.
It increases engineering effort and, more seriously, introduces the risk of making unfair comparisons.
Therefore, it is much in demand to create an \subjabbr package that can save the effort of practitioners and provide a testbed for accomplishing reproducible research.
To this end, we pinpoint what features prior frameworks lack for \subjabbr and the challenges to satisfy these requirements:

\textit{(1) Unified View for Modularized and Flexible Programming.} In each round of an FL course, a general FL algorithm (e.g., FedAvg) exchanges homogeneous data (here model parameters) for one pass. In contrast, \subjabbr algorithms~\cite{fedsage,fedgnn,gcfl} often require several heterogeneous data (e.g., gradients, node embeddings, encrypted adjacency lists, etc.) exchanges across participants. Meanwhile, besides ordinary local updates/global aggregation, participants of \subjabbr have rich kinds of subroutines to handle those heterogeneous data. FL algorithms are often expressed in most existing frameworks by declaring a static computational graph, which pushes developers to care about coordinating the participants for these data exchanges. 
Thus, an \subjabbr package should provide a unified view for developers to express such heterogeneous data exchanges and the various subroutines effortlessly, allowing flexible modularization of the rich behaviors so that \subjabbr algorithms can be implemented conveniently.

\textit{(2) Unified and Comprehensive Benchmarks.} Due to the privacy issue, real-world \subjabbr datasets are rare. Most prior \subjabbr works are evaluated by splitting a standalone graph dataset. Without a unified splitting mechanism, they essentially use their respective datasets. Meanwhile, their GNN implementations have not been aligned and integrated into the same FL framework. All these increase the risk of inconsistent comparisons of related works, urging an \subjabbr package to set up configurable, unified, and comprehensive benchmarks.

\textit{(3) Efficient and Automated Model Tuning.} Most federated optimization algorithms have not been extensively studied with GNN models. Hence, practitioners often lack proper prior for tuning their GNN models under the FL setting, making it inevitable to conduct hyper-parameter optimization (HPO). 
Moreover, directly integrating a general HPO toolkit into an FL framework cannot satisfy the efficiency requirements due to the massive cost of executing an entire FL course~\cite{fedex}. Even a single model is tuned perfectly, the prevalent non-i.i.d.ness in federated graph data might still lead to unsatisfactory performances. In this situation, monitoring the FL procedure to get aware of the non-i.i.d.ness and personalizing (hyper-)parameters are helpful for further tuning the GNN models.

\textit{(4) Privacy Attacks and Defence.} Performing privacy attacks on the FL algorithm is a direct and effective way to examine whether the FL procedure has the risk of privacy leakage. However, none of the existing FL frameworks contains this. Moreover, compared with the general FL framework, except for sharing the gradients of the global model, \subjabbr may also share additional graph-related information among clients, such as node embeddings~\cite{fedgnn} and neighbor generator~\cite{fedsage}. Without verifying the security of sharing such information, the application of \subjabbr remains questionable.

Motivated by these, in this paper, we develop an \subjabbr package \ours to satisfy these challenging requirements:

(1) We choose to build \ours upon an event-driven FL framework \ourcore~\cite{core}, which abstracts the exchanged data into messages and characterizes the behavior of each participant by defining the message handlers. Users who need to develop \subjabbr algorithms can simply define the (heterogeneous) messages and handlers, eliminating the engineering for coordinating participants. Meanwhile, different handlers can be implemented with respective graph learning backends (e.g., torch\_geometric and tf\_geometric).

(2) For the ease of benchmarking related \subjabbr methods, \ours provides a \textit{GraphDataZoo} that integrates a rich collection of splitting mechanisms applicable to most existing graph datasets and a \textit{GNNModelZoo} that integrates many state-of-the-art \subjabbr algorithms. Thus, users can reproduce the results of related works effortlessly. It is worth mentioning that we identify a unique covariate shift of graph data that comes from the graph structures, and we design a federal random graph model for the corresponding further study.

(3) \ours also provides a component for tuning the \subjabbr methods. On the one hand, it provides fundamental functionalities to achieve low-fidelity HPO, empowering users of \ours to generalize existing HPO algorithms to the FL settings. On the other hand, when a single model is inadequate to handle the non-i.i.d. graph data, our model-tuning component provides rich metrics to monitor the dissimilarity among clients and a parameter grouping mechanism for describing various personalization algorithms in a unified way.

(4) Considering the additional heterogeneous data exchanged in \subjabbr, demonstrating the level of privacy leakage under various attacks and providing effective defense strategies are indispensable. \ours includes a dedicated component to provide various off-the-shelf privacy attack and defence abilities, which are encapsulated as plug-in functions for the \subjabbr procedure.

We utilize \ours to conduct extensive experimental studies to validate the implementation correctness, verify its efficacy, and better understanding the characteristics of \subjabbr. Furthermore, we employ \ours to serve three real-world E-commerce scenarios, and the collaboratively learned GNN outperforms their locally learned counterparts, which confirms the business value of \ours. We have open-sourced \ours for the community, which we believe can ease the innovation of \subjabbr algorithms, promote their applications, and benefit more real-world business.
\vspace{-0.05in}
\section{Related Work}
\label{sec:related}

\subsection{Federated Learning}
\label{subsec:fl}
Generally, the goal of FL is to solve:
$\min_{\theta}f(\theta)=\sum_{i=1}^{N}p_i F_{i}(\theta)=\mathbb{E}[F_{i}(\theta)]$,
where $N$ is the number of clients (a.k.a. devices or parties), $F_{i}(\cdot)$ is the local objective of the $i$-th client, $p_i > 0$, and $\sum_{i}^{N}p_{i}=1$.
As a special case of distributed learning, the essential research topic for FL is its optimization approaches.
Concerning the communication cost, FedAvg~\cite{def_fl} allows clients to make more than one local update at each round.
The following works include FedProx~\cite{fedprox}, FedOPT~\cite{fedopt}, FedNOVA~\cite{fednova}, SCAFFOLD~\cite{scaffold}, etc.
These methods have been extensively studied on vision and language tasks, but when applied to optimize GNNs, their characteristics are less understood to the community.
We refer readers to the survey papers~\cite{survey0,survey1}.

\vspace{-0.1in}
\subsection{Federated Graph Learning}
\label{subsec:gfl}
When handling graph-related tasks under the FL setting, several unique algorithmic challenges emerge, e.g., complete the cross-client edges~\cite{fedsage}, handle the heterogeneous graph-level tasks~\cite{gcfl}, augment each client's subgraph~\cite{fedgnn}, and align the entities across clients~\cite{FKGE}.
Many recent \subjabbr works have attempted to resolve such challenges, which usually require exchanging heterogeneous data across the participants, and the behaviors of the participants become richer than ordinary FL methods.
These characteristics of \subjabbr algorithms lead to the unique requirements (see Sec.~\ref{sec:infra}).

\vspace{-0.1in}
\subsection{FL Software}
\label{subsec:toolkit}
With the need for FL increasing, many FL frameworks~\cite{tff,fedml,pysyft,sherpaai,pyvertical,fate,ibmfl,flower} have sprung up.
Most of them are designed as a conventional distributed machine learning framework, where a computational graph is declared and split for participants. Then each specific part is executed by the corresponding participant.
Users often have to implement their FL algorithms with declarative programming (i.e., describing the computational graph), which raises the bar for developers.
This usability issue is exacerbated in satisfying the unique requirements of \subjabbr methods.
Consequently, most existing FL frameworks have no dedicated support for \subjabbr, and practitioners cannot effortlessly build \subjabbr methods upon them.
An exception is FedML~\cite{fedml}, one of the first FL frameworks built on an event-driven architecture, provides a \subjabbr package FedGraphNN~\cite{he2021fedgraphnn}.
However, they still focus on the \subjabbr algorithms that have simple and canonical behaviors.
Many impactful \subjabbr works have not been integrated, including those discussed above.
Besides, they have ignored the requirements of efficiently tuning GNN models and conducting privacy attacks\&defence for \subjabbr algorithms, which are crucial in both practice and research.
\vspace{-0.1in}
\section{Infrastructure}
\label{sec:infra}
\begin{figure}
    \centering
    \includegraphics[width=0.45\textwidth]{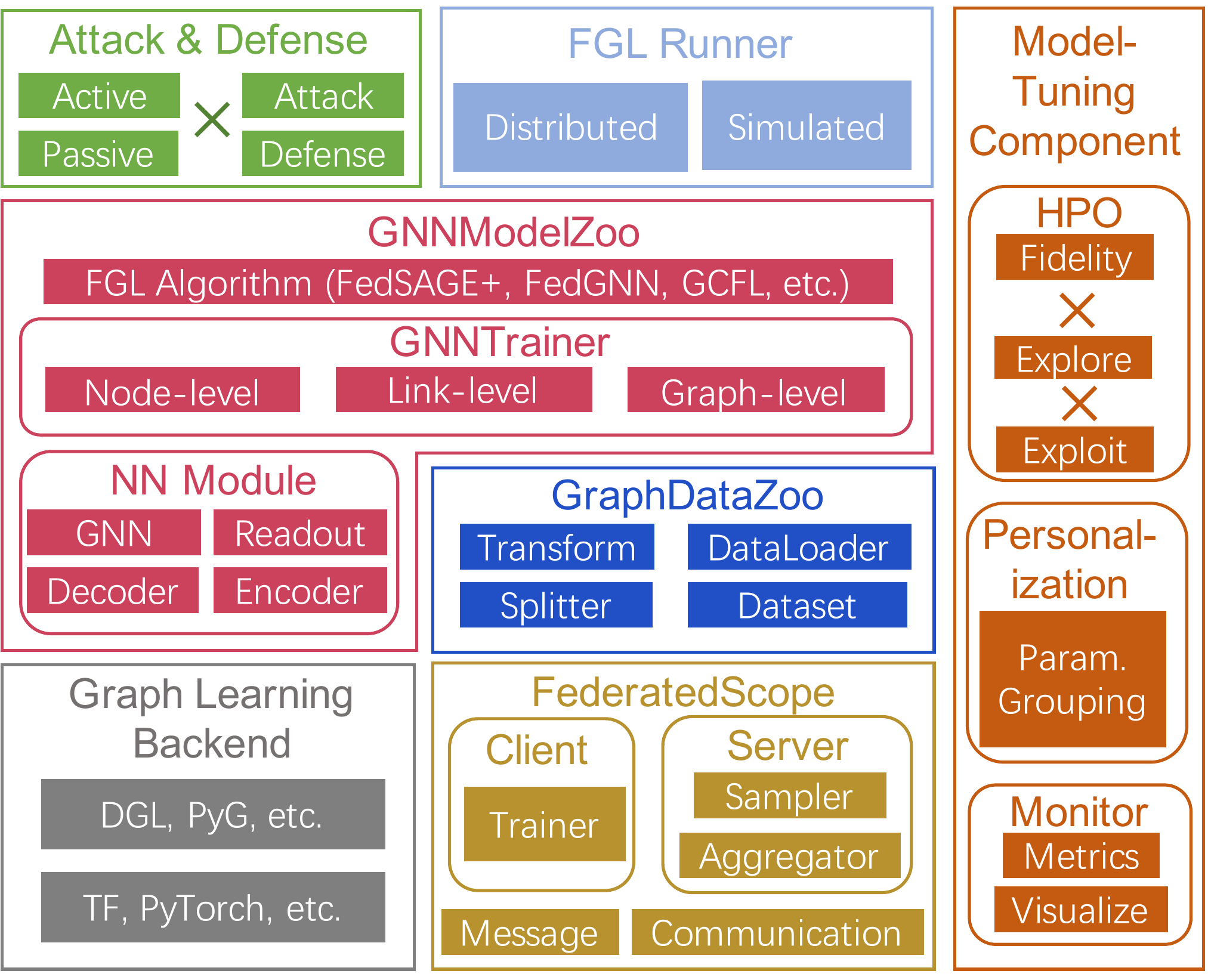}
    \vspace{-0.1in}
    \caption{Overview of \ours.}
    \label{fig:stack}
    \vspace{-0.2in}
\end{figure}
We present the overview of \ours in Fig.~\ref{fig:stack}. At the core of \ours is an event-driven FL framework \ourcore~\cite{core} with fundamental utilities (i.e., framing the FL procedure) and it is compatible with various graph learning backends. Thus, we build our \textit{GNNModelZoo} and \textit{GraphDataZoo} upon \ourcore with maximum flexibility for expressing the learning procedure and minimum care for the federal staff (e.g., coordinating participants). With such ease of development, we have integrated many state-of-the-art \subjabbr algorithms into \textit{GNNModelZoo}. We design the \textit{Runner} class as a convenient interface to access \subjabbr executions, which unifies the simulation and the distributed modes. Meanwhile, an auto model-tuning component for performance optimization and a component for privacy attack and defense purposes are provided in \ours.

\begin{table}[tp]
    \centering
    \caption{A summary of three representative \subjabbr algorithms, where we only list the behaviors other than ordinary local updates and aggregation.}
    \vspace{-0.1in}
    \label{tab:fedgl}
    \resizebox{0.475\textwidth}{!}{
    \centering
    \begin{tabular}{c|c|c|c}
    \hline
        Method & FedSage+~\cite{fedsage} & FedGNN~\cite{fedgnn} & GCFL+~\cite{gcfl} \\
        \hline
        Task & Node classification & Link prediction & Graph classification \\
        \hline
        \multirow{4}{*}{Exchange} & Model param. & Model param. & Model param. \\
        & Node emb. & Node emb. & Model grad. \\
        & NeighGen param. & Adj. list & \\
        & NeighGen grad. & & \\
        \hline
        Server & Broadcast emb. & Node clustering & Grad. clustering \\
        behavior & Broadcast grad. & Broadcast emb. & Param. deriving \\
        \hline
        Client & Send emb. and NeighGen & Generate pseudo edges & None \\
        behavior & Apply cross-client grad. & & \\
        \hline
    \end{tabular}
    }
    \vspace{-0.1in}
\end{table}
\subsection{Requirements of Federated Graph Learning}
\label{subsec:requirement}
We first review the existing \subjabbr algorithms and summarize their uniqueness against general FL. As shown in Table~\ref{tab:fedgl}, the three very recent \subjabbr works, targeting different tasks, need to exchange heterogeneous data across the participants. In contrast, in each round of a general FL procedure (e.g., using FedAvg), only homogeneous data are exchanged for one pass from server to clients and one pass back. As a result of this difference, the participant of an \subjabbr course often executes rich kinds of subroutines to handle the received data and prepare what to send. In contrast, the participant of a general FL course has canonical behaviors, i.e., local updates or aggregation.

Thus, there is a demand for a unified view to express these multiple passes of heterogeneous data exchanges and the accompanying subroutines. In this way, developers can be agnostic about the communication, better modularize the \subjabbr procedure, and choose the graph learning backend flexibly.

\vspace{-0.05in}
\subsection{Development based on FederatedScope}
\label{subsec:paradigm}
\begin{figure}[bp]
	\centering
	\vspace{-0.1in}
	\includegraphics[width=0.46\textwidth]{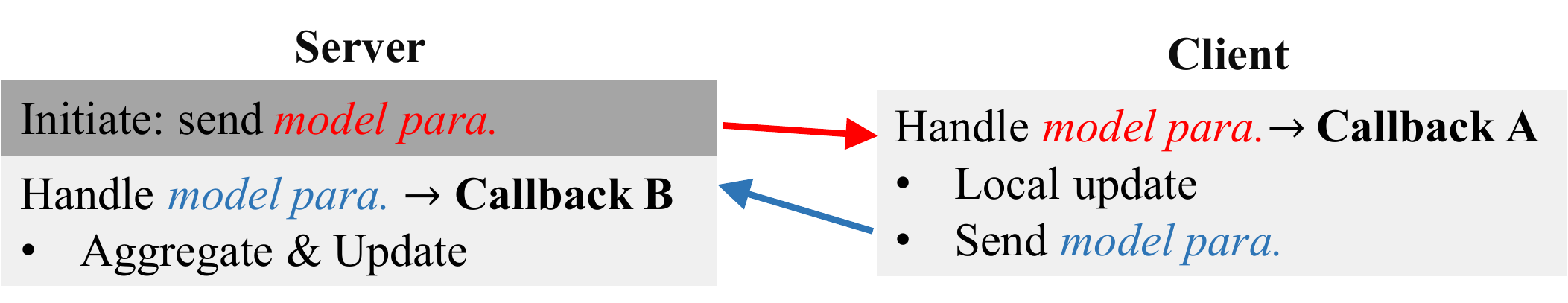}
	\vspace{-0.1in}
	\caption{Implement a standard FL algorithm based on \ourcore.~\label{fig:standard_FL}}
\end{figure}
\begin{figure*}[tb]
	\centering
	    \begin{subfigure}[t]{0.46\textwidth}
		    \centering
		    \includegraphics[width=\textwidth]{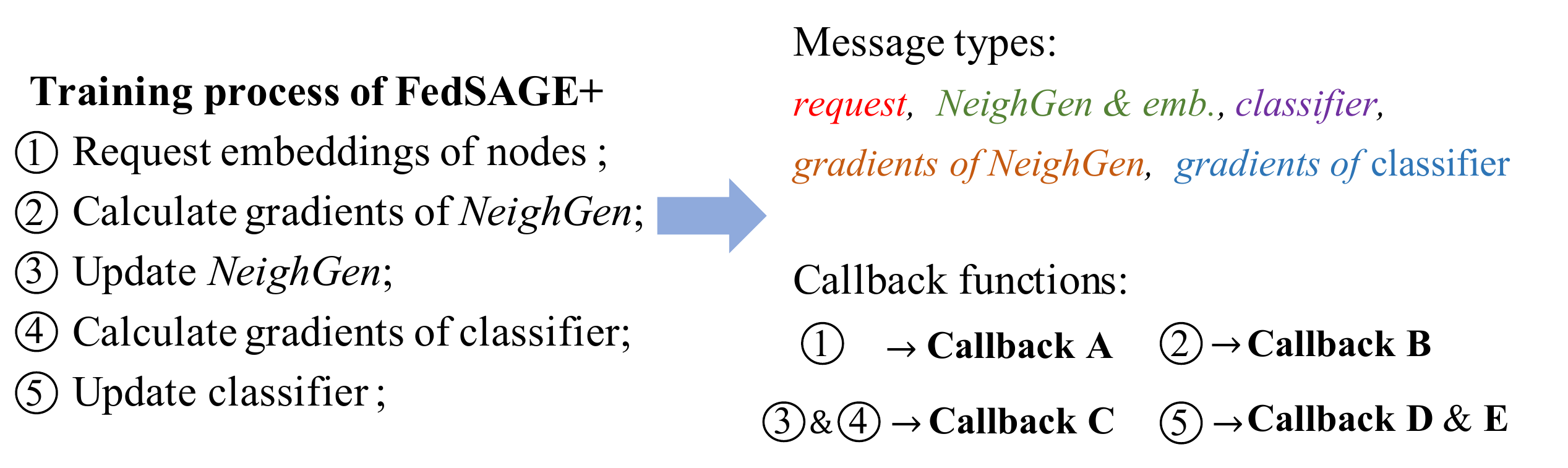}
	    \end{subfigure}
	    \begin{subfigure}[t]{0.46\textwidth}
		    \centering
		    \includegraphics[width=\textwidth]{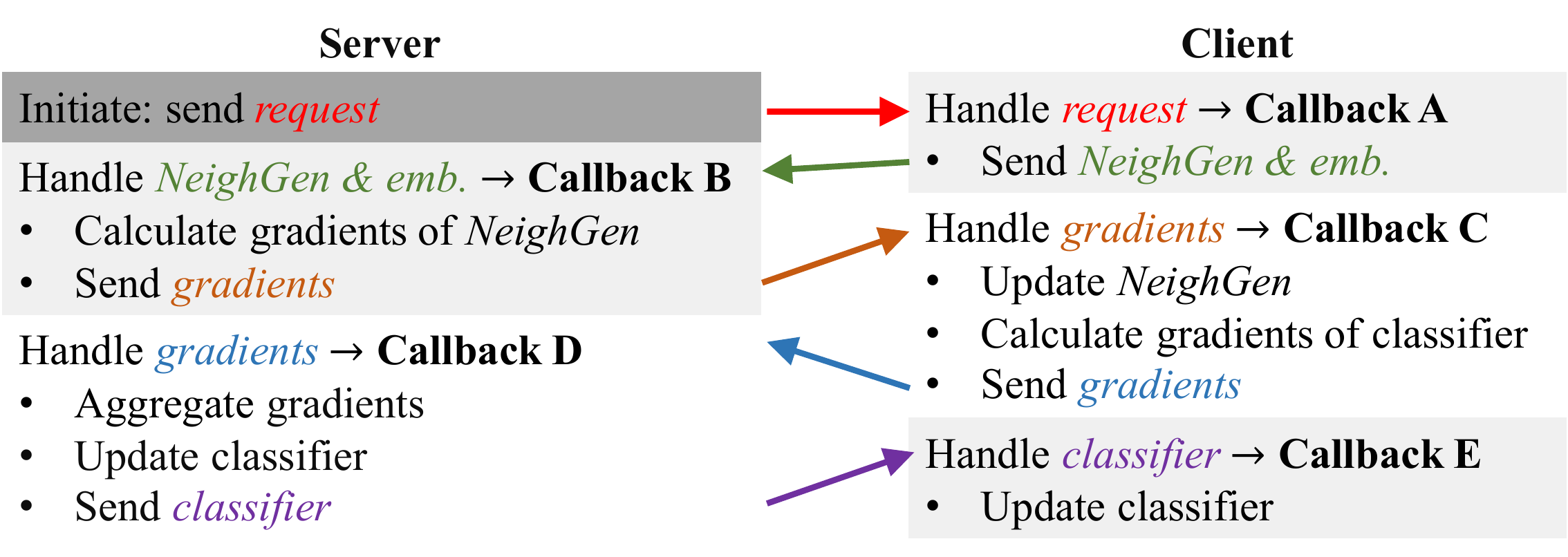}
	    \end{subfigure}
	    \caption{Implement FedSage+ based on the event-driven framework \ourcore.~\label{fig:FGL}}
\end{figure*}
To satisfy the unique requirements of \subjabbr discussed above, we develop \ours based on a event-driven FL framework named \ourcore, which abstracts the data exchange in an FL procedure as message passing.
With the help of \ourcore, implementing \subjabbr methods can be summarized in two folds: (1) defining what kinds of messages should be exchanged; (2) describing the behaviors of the server/client to handle these messages.
From such a point of view, a standard FL procedure is shown in Fig.~\ref{fig:standard_FL}, where server and client pass homogeneous messages (i.e., the model parameters).
When receiving the messages, they conduct aggregation and local updates, respectively.
As for \subjabbr algorithms, we take FedSage+ as an example.
As shown in Fig.~\ref{fig:FGL}, we extract the heterogeneous exchanged messages according to FedSage+, including model parameters, gradients, and node embeddings.
Then we frame and transform the operations defined in training steps to callback functions as subroutines to handle different types of received messages.
For example, when receiving the request, the client employs the corresponding callback function to send node embeddings and ``NeighGen'' (i.e., a neighbor generation model) back to the server.

The goal of \ours includes both \textit{convenient usage} for the existing \subjabbr methods and \textit{flexible extension} for new \subjabbr approaches.
Benefited from \ourcore, the heterogeneous exchanged data and various subroutines can be conveniently expressed as messages and handlers, which supports us to implement many state-of-the-art \subjabbr methods, including FedSage+, FedGNN, and GCFL+, by providing different kinds of message (e.g., model parameters, node embeddings, auxiliary model, adjacent list, etc) and participants' behavior (e.g., broadcast, cluster, etc).
The modularization of a whole \subjabbr procedure into messages and handlers makes it flexible for developers to express various operations defined in customized \subjabbr methods separately without considering coordinating the participants in a static computational graph.
\section{GraphDataZoo}
\label{sec:data}
A comprehensive \text{GraphDataZoo} is indispensable to provide a unified testbed for \subjabbr.
To satisfy the various experiment purposes, we allow users to constitute an FL dataset by configuring the choices of \textit{Dataset}, \textit{Splitter}, \textit{Transform}, and \textit{Dataloader}.
Conventionally, the \textit{Transform} classes are responsible for mapping each graph into another, e.g., augmenting node degree as a node attribute. The \textit{Dataloader} classes are designed for traversing a collection of graphs or sampling subgraphs from a graph. We will elaborate on the \textit{Splitter} and the \text{Dataset} classes in this section.

Tasks defined on graph data are usually categorized as follow:
\textit{(1) Node-level task:} Each instance is a node which is associated with its label. To make prediction for a node, its $k$-hop neighborhood is often considered as the input to a GNN.
\textit{(2) Link-level task:} The goal is to predict whether any given node pair is connected or the label of each given link (e.g., the rating a user assigns to an item).
\textit{(3) Graph-level task:} Each instance is an individual graph which is associated with its label.
For the link/node-level tasks, transductive setting is prevalent, where both the labeled and unlabeled links/nodes appear in the same graph. As for the graph-level task, a standalone dataset often consists of a collection of graphs.

\subsection{Splitting Standalone Datasets}
\label{subsec:split}
Existing graph datasets are a valuable source to satisfy the need for more FL datasets~\cite{survey1}.
Under the federated learning setting, the dataset is decentralized.
To simulate federated graph datasets by existing standalone ones, our \textit{GraphDataZoo} integrates a rich collection of \textit{splitters}.
These \textit{splitters} are responsible for dispersing a given standalone graph dataset into multiple clients, with configurable statistical heterogeneity among them.
For the node/link-level tasks, each client should hold a subgraph, while for the graph-level tasks, each client should hold a subset of all the graphs.

We aim to enable related works to compare on unified, configurable, and comprehensive federated graph datasets, and thus provide many off-the-shelf \textit{splitters}. Some splitters split a given dataset by specific meta data or the node attribute value, expecting to simulate realistic FL scenarios. Some other splitters are designed to provide various non-i.i.d.ness, including covariate shift, concept drift, and prior probability shift~\cite{survey1}. Details about the provided splitters and the FL datasets constructed by applying them can be found in the Appendix~\ref{subsec:splitterdetails} and Appendix~\ref{subsec:Datasets Description}, respectively.

\subsection{New Federated Learning Datasets}
\label{subsec:newdata}
In addition to the the strategy of splitting existing standalone datasets, we also construct three federated graph datasets from other real-world data sources or federal random graph model:

\textit{(1) FedDBLP}: We create this dataset from the latest DBLP dump, where each node corresponds to a published paper, and each edge corresponds to a citation. We use the bag-of-words of each paper's abstract as its node attributes and regard the theme of paper as its label. To simulate the scenario that a venue or an organizer forbids others to cite its papers, \ours allows users to split this dataset by each node's venue or the organizer of that venue.

\textit{(2) Cross-scenario recommendation (CSR)}: We create this dataset from the user-item interactions collected from an E-commerce platform. \ours allows users to split the graph by each item's category or by which scenario an interaction happens.

\textit{(3) FedcSBM}: Graph data consist of attributive and structural patterns, but prior federated graph datasets have not decoupled the covariate shifts of these two aspects. Hence, we propose a federal random graph model \textit{FedcSBM} based on cSBM~\cite{cSBM}. \textit{FedcSBM} can produce the dataset where the node attributes of different clients obey the same distribution. Meantime, the homophilic degrees can be different among the clients, so that covariate shift comes from the structural aspect.
\section{GNNModelZoo and Model-Tuning Component}
\label{sec:tune}
As an \subjabbr package, \ours provides a \textit{GNNModelZoo}.
As discussed in Sec.~\ref{sec:data}, models designated for tasks of different levels will encounter heterogeneous input and/or output, thus requiring different architectures.
Hence, in the \textit{NN module} of \ours, we modularize a general neural network model into four categories of building bricks:
\textit{(1) Encoder}: embeds the raw node attributes or edge attributes, e.g., atom encoder and bond encoder. \textit{(2) GNN}: learns discriminative representations for the nodes from their original representations (raw or encoded) and the graph structures. \textit{(3) Decoder}: recovers these hidden representations back into original node attributes or adjacency relationships. \textit{(4) Readout}: aggregates node representations into a graph representation, e.g., the mean pooling.
With a rich collection of choices in each category, users can build various kinds of neural network models out-of-the-box.
Particularly, in addition to the vanilla GNNs~\cite{GCN,chebnet,GraphSAGE,GIN,GAT}, our \textit{GNNModelZoo} also includes the GNNs that decouples feature transformation and propagation, e.g., GPR-GNN~\cite{GPRGNN}.
Such a kind of GNNs has been ignored by FedGraphNN~\cite{he2021fedgraphnn}.
However, we identify their unique advantages in \subjabbr---handling covariate shift among the client-wise graphs that comes from graph structures.

For the convenience of developers, \ours integrates the \textit{GNNTrainer} class, which encapsulates the local training procedure.
It can be easily configured to adjust for different levels of tasks and full-batch/graph-sampling settings.
Thus, developer can focus on the \subjabbr algorithms without caring for the procedure of local updates.
Then we implement many state-of-the-art \subjabbr algorithms and integrate them into our \textit{GNNModelZoo}.

With \textit{GraphDataZoo} and \textit{GNNModelZoo}, users are empowered with \subjabbr capacities.
However, the performances of \subjabbr algorithms are often sensitive to their hyper-parameters.
It is indispensable to make hyper-parameter optimization (HPO) to create reproducible research.
To this ends, \ours incorporates a \textit{model-tuning component}, which provides the functionalities for (1) making efficient HPO under the FL setting; (2) monitoring the FL procedure and making personalization to better handle non-i.i.d. data.

\vspace{-0.05in}
\subsection{Federated Hyper-parameter Optimization}
\label{subsec:hpo}
In general, HPO is a trial-and-error procedure, where, in each trial, a specific hyper-parameter configuration is proposed and evaluated.
HPO methods mainly differ from each other in exploiting the feedback of each trial and exploring the search space based on previous trials.
Whatever method, one of the most affecting issues is the cost of making an exact evaluation, which corresponds to an entire training course and then evaluation.
This issue becomes severer under the \subjabbr setting since an entire training course often consists of hundreds of communication rounds, and \subjabbr methods, in each round, often exchange additional information across participants, where even a single training course is extremely costly.

\begin{figure}[tbp]
    \centering
    \includegraphics[width=0.45\textwidth]{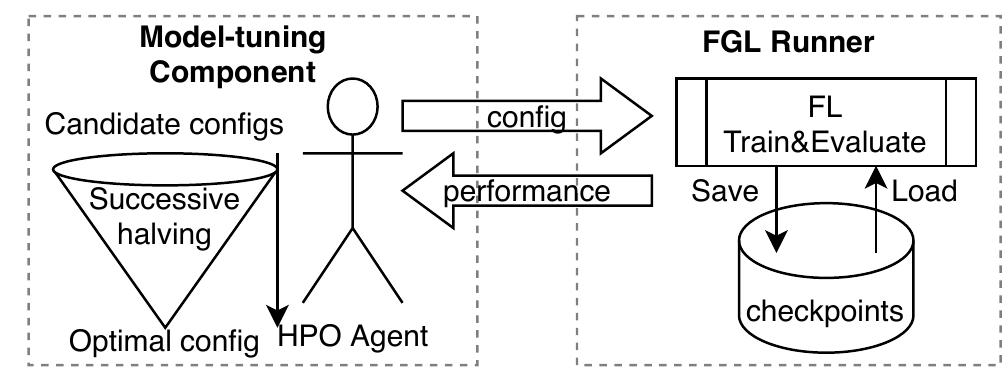}
    \vspace{-0.1in}
    \caption{An example of HPO for \subjabbr: \ours allows each training course to be restored from a given checkpoint, proceed for any specified number of rounds, and be saved for continual training.}
    \label{fig:hpo}
    \vspace{-0.1in}
\end{figure}

A general and prosperous strategy to reduce the evaluation cost is making multi-fidelity HPO~\cite{generalhpo}.
\ours allows users to reduce the fidelity by (1) making a limited number of FL rounds instead of an entire FL course for each trial; (2) sampling a small fraction of clients, e.g., $K (K\ll N)$, in each round.
We use Successive Halving Algorithm (SHA)~\cite{hyperband} as an example to show how multi-fidelity HPO methods work in the FL setting.
As Fig.~\ref{fig:hpo} shows, a set of candidate configurations are maintained, each of which will be evaluated in each stage of SHA.
When performing low-fidelity HPO, each specific configuration can be evaluated by restoring an FL course from its corresponding checkpoint (if it exists), making only a few FL rounds to update the model, and evaluating the model to acquire its performance.
Then these configurations are sorted w.r.t. their performances and only the top half of them are reserved for the next stage of SHA.
This procedure continues until only one configuration remains, regarded as the optimal one.

We identify the requirement of functionalities to save and restore an \subjabbr training course from this algorithmic example. These functionalities are also indispensable to achieving a reliable failover mechanism. Thus, we first show which factors determine the state of an FL course: \textit{(1) Server-side}: Basically, the current round number and the global model parameters must be saved. When the aggregator is stateful, e.g., considering momentum, its maintained state, e.g., the moving average of a certain quantity, needs to be kept. \textit{(2) Client-side}: When the local updates are made with mini-batch training, the client-specific data-loader is usually stateful, whose index and order might need to be held. When personalization is utilized, the client-specific model parameters need to be saved.
With sufficient factors saved as a checkpoint, \ours can restore an \subjabbr training course from it and proceed.

Meanwhile, we follow the design of \ourcore and make the entry interface of \ours as a callable \subjabbr runner, which receives a configuration and returns a collection of metrics for the conducted \subjabbr training course (entire or not). Consequently, each HPO algorithm incorporated in our model-tuning component can be abstracted as an agent, repeatedly calling the \subjabbr runner to collect the feedback. Benefiting from this interface design and the capacity to save and restore an \subjabbr training course, any one-shot HPO method can be effortlessly generalized to the \subjabbr setting upon \ours.
It is worth mentioning that what to return by the \subjabbr runner is configurable, where efficiency-related metrics, e.g., the average latency of each round, can be included. Therefore, optimizing hyper-parameters from the system perspective~\cite{fedhposys} is also supported by \ours.

\subsection{Monitoring and Personalization}
\label{subsec:personalization}
\begin{figure}[tbp]
    \centering
    \begin{subfigure}[b]{0.22\textwidth}
    \centering
    \includegraphics[width=\textwidth]{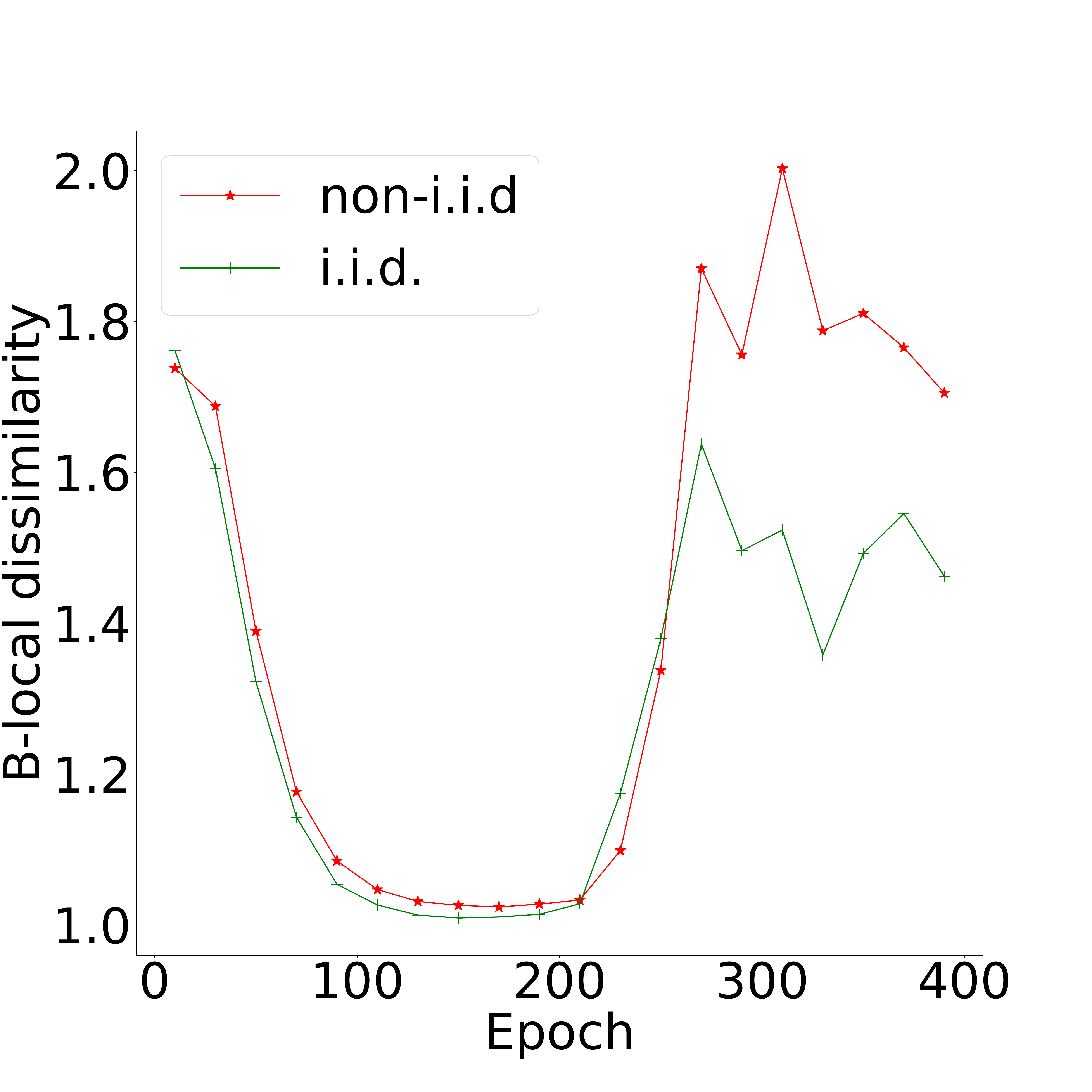}
    \caption{Monitoring the \subjabbr course on i.i.d. and non-i.i.d. datasets constructed by \textit{FedcSBM}.}
    \label{subfig:dissim}
    \end{subfigure}
    \hfill
    \begin{subfigure}[b]{0.22\textwidth}
    \centering
    \includegraphics[width=\textwidth]{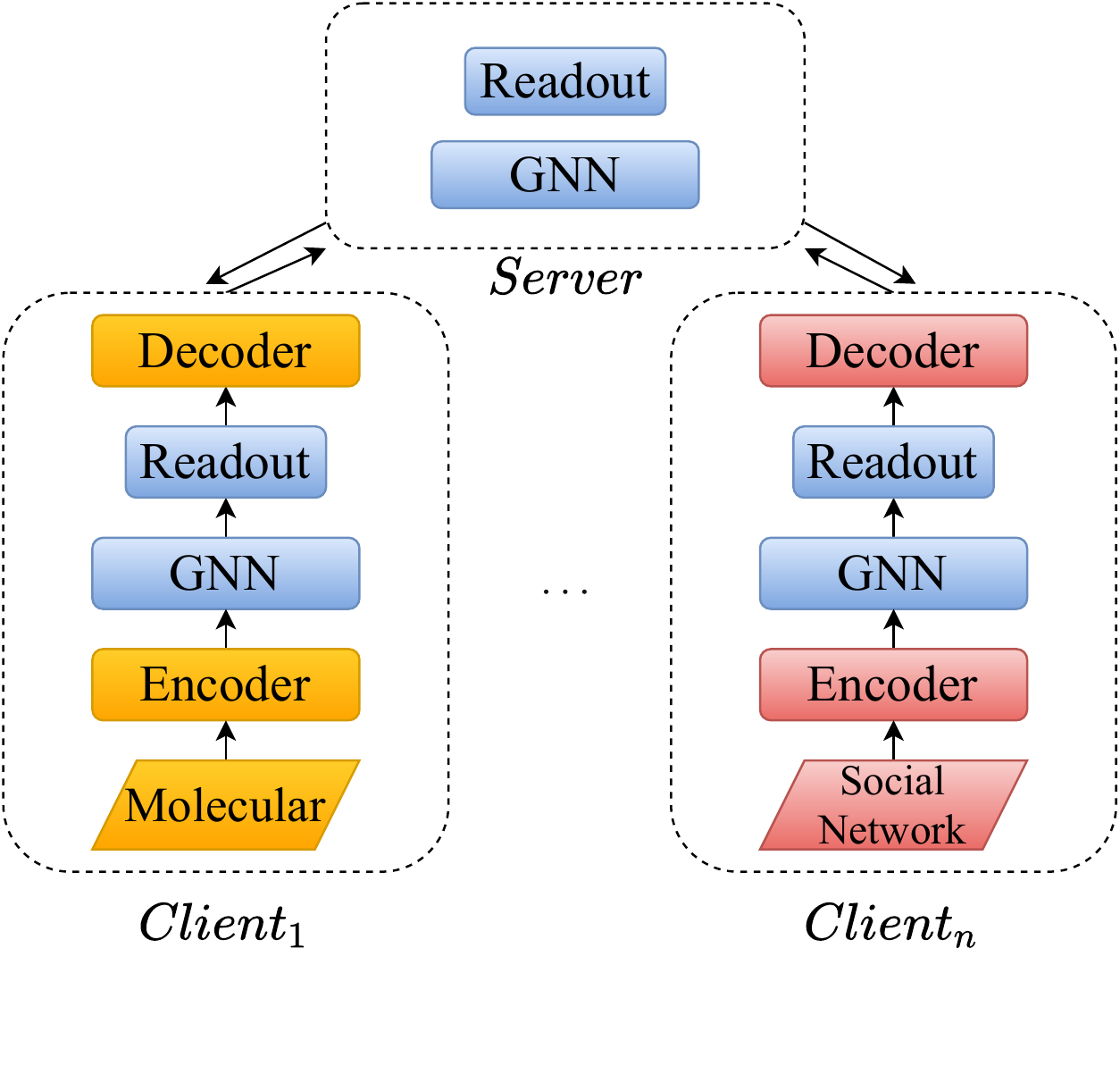}
    \caption{An example of personalizing GNN: Each client has its dedicated encoder and decoder.}
    \label{subfig:psn}
    \end{subfigure}
    \vspace{-0.1in}
    \caption{Examples of monitoring and personalization.}
    \vspace{-0.1in}
    \label{fig:monandpsn}
\end{figure}

Practitioners often monitor the learning procedure by visualizing the curves of training loss and validation performance to understand whether the learning has converged and the GNN model has overfitted. When it comes to \subjabbr, we consider both the client-wise metrics, e.g., the local training loss, and some metrics to be calculated at the server-side. Specifically, we have implemented several metrics, including B-local dissimilarity~\cite{fedprox} and the covariance matrix of gradients, calculated from the aggregated messages to reflect the statistical heterogeneity among clients. The larger these metrics are, the more different client-wise graphs are. As shown in Fig.~\ref{subfig:dissim}, the B-local dissimilarity on non-i.i.d. data is larger than that on i.i.d. data at almost all stages of the training course, which becomes particularly noticeable at the end.

Then we build the monitoring functionality upon related toolkits (e.g., WandB and TensorBoard) to log and visualize the metrics. To use the out-off-shelf metrics, users only need to specify them in the configuration. Meantime, \ours has provided the API for users to register any quantity calculated/estimated during the local update/aggregation, which would be monitored in the execution.

\begin{table*}[htbp]
\centering
\small
\caption{Results on representative node classification datasets with \textit{random\_splitter}: Mean accuracy $\pm$ standard deviation.}
\label{tab:nodelevel_rand}
\resizebox{\textwidth}{!}{
\begin{tabular}{c|ccccc|ccccc|ccccc}
\hline
 &
  \multicolumn{5}{c|}{Cora} &
  \multicolumn{5}{c|}{CiteSeer} &
  \multicolumn{5}{c}{PubMed} \\
 &
  Local &
  FedAvg &
  FedOpt &
  FedProx &
  Global &
  Local &
  FedAvg &
  FedOpt &
  FedProx &
  Global &
  Local &
  FedAvg &
  FedOpt &
  FedProx &
  Global \\ \hline
GCN &
  \multicolumn{1}{c|}{80.95±1.49} &
  \multicolumn{1}{c|}{86.63±1.35} &
  \multicolumn{1}{c|}{86.11±1.29} &
  \multicolumn{1}{c|}{86.60±1.59} &
  86.89±1.82 &
  \multicolumn{1}{c|}{74.29±1.35} &
  \multicolumn{1}{c|}{76.48±1.52} &
  \multicolumn{1}{c|}{77.43±0.90} &
  \multicolumn{1}{c|}{77.29±1.20} &
  77.42±1.15 &
  \multicolumn{1}{c|}{85.25±0.73} &
  \multicolumn{1}{c|}{85.29±0.95} &
  \multicolumn{1}{c|}{84.39±1.53} &
  \multicolumn{1}{c|}{85.21±1.17} &
  85.38±0.33 \\
GraphSAGE &
  \multicolumn{1}{c|}{75.12±1.54} &
  \multicolumn{1}{c|}{85.42±1.80} &
  \multicolumn{1}{c|}{84.73±1.58} &
  \multicolumn{1}{c|}{84.83±1.66} &
  86.86±2.15 &
  \multicolumn{1}{c|}{73.30±1.30} &
  \multicolumn{1}{c|}{76.86±1.38} &
  \multicolumn{1}{c|}{75.99±1.96} &
  \multicolumn{1}{c|}{78.05±0.81} &
  77.48±1.27 &
  \multicolumn{1}{c|}{84.58±0.41} &
  \multicolumn{1}{c|}{86.45±0.43} &
  \multicolumn{1}{c|}{85.67±0.45} &
  \multicolumn{1}{c|}{86.51±0.37} &
  86.23±0.58 \\
GAT &
  \multicolumn{1}{c|}{78.86±2.25} &
  \multicolumn{1}{c|}{85.35±2.29} &
  \multicolumn{1}{c|}{84.40±2.70} &
  \multicolumn{1}{c|}{84.50±2.74} &
  85.78±2.43 &
  \multicolumn{1}{c|}{73.85±1.00} &
  \multicolumn{1}{c|}{76.37±1.11} &
  \multicolumn{1}{c|}{76.96±1.75} &
  \multicolumn{1}{c|}{77.15±1.54} &
  76.91±1.02 &
  \multicolumn{1}{c|}{83.81±0.69} &
  \multicolumn{1}{c|}{84.66±0.74} &
  \multicolumn{1}{c|}{83.78±1.11} &
  \multicolumn{1}{c|}{83.79±0.87} &
  84.89±0.34 \\
GPR-GNN &
  \multicolumn{1}{c|}{84.90±1.13} &
  \multicolumn{1}{c|}{89.00±0.66} &
  \multicolumn{1}{c|}{87.62±1.20} &
  \multicolumn{1}{c|}{88.44±0.75} &
  88.54±1.58 &
  \multicolumn{1}{c|}{74.81±1.43} &
  \multicolumn{1}{c|}{79.67±1.41} &
  \multicolumn{1}{c|}{77.99±1.25} &
  \multicolumn{1}{c|}{79.35±1.11} &
  79.67±1.42 &
  \multicolumn{1}{c|}{86.85±0.39} &
  \multicolumn{1}{c|}{85.88±1.24} &
  \multicolumn{1}{c|}{84.57±0.68} &
  \multicolumn{1}{c|}{86.92±1.25} &
  85.15±0.76 \\
  \hline
\end{tabular}
}
\end{table*}

\begin{table*}[htbp]
\centering
\small
\caption{Results on representative node classification datasets with \textit{community\_splitter}: Mean accuracy $\pm$ standard deviation.}
\label{tab:nodelevel_com}
\resizebox{\textwidth}{!}{
\begin{tabular}{c|ccccc|ccccc|ccccc}
\hline
 &
  \multicolumn{5}{c|}{Cora} &
  \multicolumn{5}{c|}{CiteSeer} &
  \multicolumn{5}{c}{PubMed} \\
 &
  Local &
  FedAvg &
  FedOpt &
  FedProx &
  Global &
  Local &
  FedAvg &
  FedOpt &
  FedProx &
  Global &
  Local &
  FedAvg &
  FedOpt &
  FedProx &
  Global \\ \hline
GCN &
  \multicolumn{1}{c|}{65.08±2.39} &
  \multicolumn{1}{c|}{87.32±1.49} &
  \multicolumn{1}{c|}{87.29±1.65} &
  \multicolumn{1}{c|}{87±16±1.51} &
  86.89±1.82 &
  \multicolumn{1}{c|}{67.53±1.87} &
  \multicolumn{1}{c|}{77.56±1.45} &
  \multicolumn{1}{c|}{77.80±0.99} &
  \multicolumn{1}{c|}{77.62±1.42} &
  77.42±1.15 &
  \multicolumn{1}{c|}{77.01±3.37} &
  \multicolumn{1}{c|}{85.24±0.69} &
  \multicolumn{1}{c|}{84.11±0.87} &
  \multicolumn{1}{c|}{85.14, 0.88} &
  85.38±0.33 \\
GraphSAGE &
  \multicolumn{1}{c|}{61.29±3.05} &
  \multicolumn{1}{c|}{87.19±1.28} &
  \multicolumn{1}{c|}{87.13±1.47} &
  \multicolumn{1}{c|}{87.09±1.46} &
  86.86±2.15 &
  \multicolumn{1}{c|}{66.17±1.50} &
  \multicolumn{1}{c|}{77.80±1.03} &
  \multicolumn{1}{c|}{78.54±1.05} &
  \multicolumn{1}{c|}{77.70±1.09} &
  77.48±1.27 &
  \multicolumn{1}{c|}{78.35±2.15} &
  \multicolumn{1}{c|}{86.87±0.53} &
  \multicolumn{1}{c|}{85.72±0.58} &
  \multicolumn{1}{c|}{86.65±0.60} &
  86.23±0.58 \\
GAT &
  \multicolumn{1}{c|}{61.53±2.81} &
  \multicolumn{1}{c|}{86.08±2.52} &
  \multicolumn{1}{c|}{85.65±2.36} &
  \multicolumn{1}{c|}{85.68±2.68} &
  85.78±2.43 &
  \multicolumn{1}{c|}{66.17±1.31} &
  \multicolumn{1}{c|}{77.21±0.97} &
  \multicolumn{1}{c|}{77.34±1.33} &
  \multicolumn{1}{c|}{77.26±1.02} &
  76.91±1.02 &
  \multicolumn{1}{c|}{75.97±3.32} &
  \multicolumn{1}{c|}{84.38±0.82} &
  \multicolumn{1}{c|}{83.34±0.87} &
  \multicolumn{1}{c|}{84.34±0.63} &
  84.89±0.34 \\
GPR-GNN &
  \multicolumn{1}{c|}{69.32±2.07} &
  \multicolumn{1}{c|}{88.93±1.64} &
  \multicolumn{1}{c|}{88.37±2.12} &
  \multicolumn{1}{c|}{88.80±1.29} &
  88.54±1.58 &
  \multicolumn{1}{c|}{71.30±1.65} &
  \multicolumn{1}{c|}{80.27±1.28} &
  \multicolumn{1}{c|}{78.32±1.45} &
  \multicolumn{1}{c|}{79.73±1.52} &
  79.67±1.42 &
  \multicolumn{1}{c|}{78.52±3.61} &
  \multicolumn{1}{c|}{85.06±0.82} &
  \multicolumn{1}{c|}{84.30±1.57} &
  \multicolumn{1}{c|}{86.77±1.16} &
  85.15±0.76 \\
  \hline
\end{tabular}
}
\end{table*}

Once some monitored metrics indicate the existence of non-i.i.d.ness, users can further tune their GNN by personalizing the model parameters and even the hyper-parameters. We present an example of personalization in Fig.~\ref{subfig:psn}. Each client has its specific encoder and decoder as the tasks among the clients come from different domains with different node attributes and node classes. In practice, a more fine-grained personalization might be preferred, where only some layers or even some variables are client-specific.

To satisfy such purposes, \ours first allows users to instantiate the model of each participant individually. Then we build a flexible parameter grouping mechanism upon the naming systems of underlying machine learning engines. Specifically, this mechanism allows users to easily declare each part (with flexible granularity) of the model as client-specific or shared. Only the shared parts will be exchanged.
\section{Off-the-shelf Attack and Defence Abilities}
\label{sec:attdef}
The privacy attack\&defence component of \ourcore has integrated various off-the-shelf passive privacy attack methods, including class representative inference attack, membership inference attack, property inference attack, and training inputs and labels inference attack. These methods have been encapsulated as optional hooks for our \textit{GNNTrainer}. Once the user has selected a specific hook, the GNN model and some needed information about the target data would automatically be fed into the hook during the training procedure. Besides the previous passive attack setting where the adversaries are honest-but-curious, \ours also supports the malicious adversary setting. The attackers can deviate from the FL protocol by modifying the messages.
To defend the passive privacy attacks, \ours can leverage \ourcore's plug-in defense strategies, including differential privacy, MPC, and data compression. Meanwhile, \ourcore provides the information checking mechanism to effectively detect anomaly messages and defend the malicious attacks.
\section{Experiments}
\label{sec:exp}
We utilize \ours to conduct extensive experiments, with the aim to validate the implementation correctness of \ours, set up benchmarks for \subjabbr that have long been demanded, and gain more insights about \subjabbr. Furthermore, we deploy \ours in real-world E-commerce scenarios to evaluate its business value.

\vspace{-0.05in}
\subsection{An Extensive Study about Federated Graph Learning}
\label{subsec:empirical}
In this study, we consider three different settings: (1) \textsc{Local}: Each client trains a GNN model with its data. (2) \subjabbr: FedAvg~\cite{def_fl}, FedOpt~\cite{fedopt}, and FedProx~\cite{fedprox} are applied to collaboratively train a GNN model on the dispersed data, respectively. (3) \textsc{Global}: One GNN model is trained on the completed dataset. By comparing these settings with various GNN architectures and on diverse tasks, we intend to set up comprehensive and solid benchmarks for \subjabbr.

\subsubsection{Node-level Tasks}\hfill\\
\label{subsubsec:node-level-tasks}
\noindent\textbf{Protocol}. For the purposes of training and validation, the client-wise data are subgraphs deduced from the original graph. Yet, global evaluation is considered for testing, where models are evaluated on the test nodes of original graph.
Thus, in both local and \subjabbr settings, we train and validate each client's model on its incomplete subgraph and test the model on the complete global graph.
In the global setting, we train and evaluate each model on the complete global graph.
For each setting, we consider popular GNN architectures: GCN, GraphSage, GAT, and GPR-GNN.
To conduct the experiments uniformly and fairly, we split the nodes into train/valid/test sets, where the ratio is $60\%:20\%:20\%$ for citation networks and $50\%:20\%:30\%$ for FedDBLP.
We randomly generate five splits for each dataset.
Each model is trained and evaluated with these five splits, and we report the averaged metric and the standard deviation.
To compare the performance of each model, we choose accuracy as the metric for all node-level tasks. In addition, we perform hyper-parameter optimization (HPO) for all methods with the learning rate $\in\{0.01, 0.05, 0.25\}$ in all settings, and the local update steps $\in\{1, 4, 16\}$ in \subjabbr.

\begin{table*}[htbp]
\centering
\caption{Results on representative link prediction datasets with \textit{label\_space\_splitter}: Hits@$n$.}
\label{tab:linklevel_kg}
\resizebox{\textwidth}{!}{
\begin{tabular}{c|ccccccccccccccc|ccccccccccccccc}
\hline
 &
  \multicolumn{15}{c|}{WN18} &
  \multicolumn{15}{c}{FB15k-237} \\
 &
  \multicolumn{3}{c}{Local} &
  \multicolumn{3}{c}{FedAvg} &
  \multicolumn{3}{c}{FedOpt} &
  \multicolumn{3}{c}{FedProx} &
  \multicolumn{3}{c|}{Global} &
  \multicolumn{3}{c}{Local} &
  \multicolumn{3}{c}{FedAvg} &
  \multicolumn{3}{c}{FedOpt} &
  \multicolumn{3}{c}{FedProx} &
  \multicolumn{3}{c}{Global} \\ \cline{2-31} 
 &
  1 &
  5 &
  \multicolumn{1}{c|}{10} &
  1 &
  5 &
  \multicolumn{1}{c|}{10} &
  1 &
  5 &
  \multicolumn{1}{c|}{10} &
  1 &
  5 &
  \multicolumn{1}{c|}{10} &
  1 &
  5 &
  10 &
  1 &
  5 &
  \multicolumn{1}{c|}{10} &
  1 &
  5 &
  \multicolumn{1}{c|}{10} &
  1 &
  5 &
  \multicolumn{1}{c|}{10} &
  1 &
  5 &
  \multicolumn{1}{c|}{10} &
  1 &
  5 &
  10 \\ \hline
GCN &
  20.70 &
  55.34 &
  \multicolumn{1}{c|}{73.85} &
  30.00 &
  79.72 &
  \multicolumn{1}{c|}{96.67} &
  22.13 &
  78.96 &
  \multicolumn{1}{c|}{94.07} &
  27.32 &
  83.01 &
  \multicolumn{1}{c|}{96.38} &
  29.67 &
  86.73 &
  97.05 &
  6.07 &
  20.29 &
  \multicolumn{1}{c|}{30.35} &
  9.86 &
  34.27 &
  \multicolumn{1}{c|}{48.02} &
  4.12 &
  18.07 &
  \multicolumn{1}{c|}{31.79} &
  4.66 &
  28.74 &
  \multicolumn{1}{c|}{41.67} &
  7.80 &
  32.46 &
  44.64 \\
GraphSAGE &
  21.06 &
  54.12 &
  \multicolumn{1}{c|}{79.88} &
  23.14 &
  78.85 &
  \multicolumn{1}{c|}{93.70} &
  22.82 &
  79.86 &
  \multicolumn{1}{c|}{93.12} &
  23.14 &
  78.52 &
  \multicolumn{1}{c|}{93.67} &
  24.24 &
  79.86 &
  93.84 &
  3.95 &
  14.64 &
  \multicolumn{1}{c|}{24.47} &
  7.13 &
  23.38 &
  \multicolumn{1}{c|}{36.60} &
  2.20 &
  19.21 &
  \multicolumn{1}{c|}{27.64} &
  5.85 &
  24.05 &
  \multicolumn{1}{c|}{36.33} &
  6.19 &
  23.57 &
  35.98 \\
GAT &
  20.89 &
  49.42 &
  \multicolumn{1}{c|}{72.48} &
  23.14 &
  77.62 &
  \multicolumn{1}{c|}{93.49} &
  23.14 &
  74.64 &
  \multicolumn{1}{c|}{93.52} &
  23.53 &
  78.40 &
  \multicolumn{1}{c|}{93.00} &
  24.24 &
  80.18 &
  93.76 &
  3.44 &
  15.02 &
  \multicolumn{1}{c|}{25.14} &
  6.06 &
  25.76 &
  \multicolumn{1}{c|}{39.04} &
  2.71 &
  18.89 &
  \multicolumn{1}{c|}{32.76} &
  6.19 &
  25.09 &
  \multicolumn{1}{c|}{38.00} &
  6.94 &
  24.43 &
  37.87 \\
GPR-GNN &
  22.86 &
  60.45 &
  \multicolumn{1}{c|}{80.73} &
  26.67 &
  82.35 &
  \multicolumn{1}{c|}{96.18} &
  24.46 &
  73.33 &
  \multicolumn{1}{c|}{87.18} &
  27.62 &
  81.87 &
  \multicolumn{1}{c|}{95.68} &
  29.19 &
  82.34 &
  96.24 &
  4.45 &
  13.26 &
  \multicolumn{1}{c|}{21.24} &
  9.62 &
  32.76 &
  \multicolumn{1}{c|}{45.97} &
  2.01 &
  9.81 &
  \multicolumn{1}{c|}{16.65} &
  3.72 &
  15.62 &
  \multicolumn{1}{c|}{27.79} &
  10.62 &
  33.87 &
  47.45 \\ \hline
\end{tabular}
}
\end{table*}

\begin{table*}[htbp]
\centering
\small
\caption{Results on representative graph classification datasets: Mean accuracy (\%) ± standard deviation.}
\label{tab:graphlevel}
\resizebox{\textwidth}{!}{
\begin{tabular}{c|ccccc|ccccc|ccccc}
\hline
 &
  \multicolumn{5}{c|}{PROTEINS} &
  \multicolumn{5}{c|}{IMDB} &
  \multicolumn{5}{c}{Multi-task} \\
 &
  Local &
  FedAvg &
  FedOpt &
  FedProx &
  Global &
  Local &
  FedAvg &
  FedOpt &
  FedProx &
  Global &
  Local &
  FedAvg &
  FedOpt &
  FedProx &
  Global \\ \hline
GCN &
  \multicolumn{1}{c|}{71.10±4.65} &
  \multicolumn{1}{c|}{73.54±4.48} &
  \multicolumn{1}{c|}{71.24±4.17} &
  \multicolumn{1}{c|}{73.36±4.49} &
  71.77±3.62 &
  \multicolumn{1}{c|}{50.76±1.14} &
  \multicolumn{1}{c|}{53.24±6.04} &
  \multicolumn{1}{c|}{50.49±8.32} &
  \multicolumn{1}{c|}{48.72±6.73} &
  53.24±6.04 &
  \multicolumn{1}{c|}{66.37±1.78} &
  \multicolumn{1}{c|}{65.99±1.18} &
  \multicolumn{1}{c|}{69.10±1.58} &
  \multicolumn{1}{c|}{68.59±1.99} &
  - \\
GIN &
  \multicolumn{1}{c|}{69.06±3.47} &
  \multicolumn{1}{c|}{73.74±5.71} &
  \multicolumn{1}{c|}{60.14±1.22} &
  \multicolumn{1}{c|}{73.18±5.66} &
  72.47±5.53 &
  \multicolumn{1}{c|}{55.82±7.56} &
  \multicolumn{1}{c|}{64.79±10.55} &
  \multicolumn{1}{c|}{51.87±6.82} &
  \multicolumn{1}{c|}{70.65±8.35} &
  72.61±2.44 &
  \multicolumn{1}{c|}{75.05±1.81} &
  \multicolumn{1}{c|}{63.40±2.22} &
  \multicolumn{1}{c|}{63.33±1.18} &
  \multicolumn{1}{c|}{63.01±0.44} &
  - \\
GAT &
  \multicolumn{1}{c|}{70.75±3.33} &
  \multicolumn{1}{c|}{71.95±4.45} &
  \multicolumn{1}{c|}{71.07±3.45} &
  \multicolumn{1}{c|}{72.13±4.68} &
  72.48±4.32 &
  \multicolumn{1}{c|}{53.12±5.81} &
  \multicolumn{1}{c|}{53.24±6.04} &
  \multicolumn{1}{c|}{47.94±6.53} &
  \multicolumn{1}{c|}{53.82±5.69} &
  53.24±6.04 &
  \multicolumn{1}{c|}{67.72±3.48} &
  \multicolumn{1}{c|}{66.75±2.97} &
  \multicolumn{1}{c|}{69.58±1.21} &
  \multicolumn{1}{c|}{69.65±1.14} &
  - \\
  \hline
\end{tabular}
}
\vspace{-0.1in}
\end{table*}

\noindent\textbf{Results and Analysis}. We present the results on three citation networks in Table~\ref{tab:nodelevel_rand} and Table~\ref{tab:nodelevel_com}, with \textit{random\_splitter} and \textit{community\_splitter}, respectively. Overall, \subjabbr can improve performance regarding those individually trained on local data, as the experience in vision and language domains suggests. In most cases on the randomly split datasets, the GNN trained on the global (original) data performs better than that of \subjabbr. With the splitting, the union of all client-wise graphs still has fewer edges than the original graph, which may explain this performance gap. In contrast, the \subjabbr setting often performs better than those trained on the global graph when our \textit{community\_splitter} constructs the federated graph datasets. At first glance, this phenomenon is counterintuitive, as the splitter also removes some of the original edges. However, these citation graphs exhibit homophily, saying that nodes are more likely to connect to nodes with the same label than those from other classes. When split by the community detection-based algorithm, we identify the changes that removed edges often connect nodes with different labels, which improves the homophilic degree of the graphs. As a result, the resulting federated graph dataset becomes easier to handle by most GNN architecture. Studies on FedSage+ and the comparisons on \textit{FedDBLP} are deferred to Appendix~\ref{subsec:expdetail}.

\vspace{-0.05in}
\subsubsection{Link-level Tasks}\hfill\\
\label{subsubsec:link-level-tasks}
\noindent\textbf{Protocol}. We largely follow the experimental setup in Sec.~\ref{subsubsec:node-level-tasks}. Specifically, we split the links into train/valid/test sets, where the ratio is $80\%:10\%:10\%$ for recommendation dataset Ciao. We use the official train/valid/test split for the knowledge graphs WN18 and FB15k-237. In addition, the link predictor for each model is a two-layer MLP with 64 as the hidden layer dimension, where the input is the element-wise multiplication of the node embeddings of the node pair. We report the mean accuracy for the recommendation dataset and Hits at 1, 5, and 10 for the knowledge graphs.

\noindent\textbf{Results and Analysis}. We present the results on the knowledge graphs in Table~\ref{tab:linklevel_kg} and the results on \textit{Ciao} in Table~\ref{tab:linklevel_rec}. Overall, the performances of the \subjabbr setting are better than those of the local setting, which is consistent with the conclusion drawn from node-level tasks and further confirms the effectiveness of \subjabbr. Notably, the splitting strategy results in unbalanced relation distributions among the clients on knowledge graphs, making the triplets of each client insufficient to characterize the entities. Thus, the performance gaps between the local and the \subjabbr settings are broad. Meantime, the performances of the \subjabbr setting become comparable to those of the global setting.

\subsubsection{Graph-level Tasks}\hfill\\
\label{subsubsec:graph-level-tasks}
\noindent\textbf{Protocol}. We largely follow the experimental setup in Sec.~\ref{subsubsec:link-level-tasks}. Specifically, a pre-linear layer is added to encode the categorical attributes preceding GNN. And a two-layer MLP is added to adapt GNN to multi-task classification as a classifier following GNN. Notably, for multi-task \subjabbr in multi-domain datasets, only the parameters of GNN are shared.

\noindent\textbf{Results and Analysis}. We present the main results in Table~\ref{tab:graphlevel}, where there is no global setting on \textit{Multi-task} dataset, as the same model cannot handle the graphs of different tasks. Overall, the relationships between different settings have been preserved compared to those on node-level tasks. One exception is on the \textit{Multi-task} dataset, where the non-i.i.d.ness is severe, e.g., the average number of nodes of the graphs on a client deviates from other clients a lot. Although we have personalized the models by \ours, the gradients/updates collected at the server-side might still have many interferences. Without dedicated techniques to handle these, e.g., GCFL+, FedAvg cannot surpass the individually trained models.
But with federated optimization algorithms, such as FedOpt and FedProx, some GNN architectures surpass the individually trained models, which proves the effectiveness of these federated optimization algorithms.
In addition, we implement personalized FL algorithms (here FedBN~\cite{fedbn} and Ditto~\cite{ditto}) and apply them for GIN on the multi-task dataset.
FedBN and Ditto lead to performances (i.e., mean accuracy with standard deviation) 72.90±1.33 (\%) and  63.35±0.6 (\%), respectively.
Due to the label unbalance, all kinds of GNNs result in the same accuracy on the \textit{HIV} dataset. Hence, we have to solely report their ROC-AUC in Table~\ref{tab:graphlevel_hiv} of Appendix~\ref{subsec:expdetail}, from which we can draw similar conclusions. Meantime, GIN surpasses GCN with FedAvg on almost all datasets, which implies the capability of distinguishing isomorphism graphs is still critical for federated graph-level tasks.
Our studies about GCFL+ are deferred to Appendix~\ref{subsec:expdetail}.

\subsection{Study about Hyper-parameter Optimization}
\label{subsec:hpo_exp}
In this experiment, we conduct HPO for federated learned GCN and GPR-GNN, intending to verify the effectiveness of our model-tuning component. Moreover, we can set up an HPO benchmark for the \subjabbr setting and attain some insights, which have long been missed but strongly demanded.

\noindent\textbf{Protocol}. We take SHA (see Sec.~\ref{subsec:hpo}) to optimize the hyper-parameters for GCN/GPR-GNN learned by FedAvg. we study low-fidelity HPO, where the number of FL rounds for each evaluation $\in\{1, 2, 4, 8\}$, and the client sampling rate for each round of FedAvg $\in\{20\%, 40\%, 80\%, 100\%\}$.
For each choice of fidelity, we repeat the SHA five times with different seeds, and we report the average ranking of its searched hyper-parameter configuration.

\begin{figure}[htbp]
    \centering
    \begin{subfigure}[b]{0.22\textwidth}
    \centering
    \includegraphics[width=\textwidth]{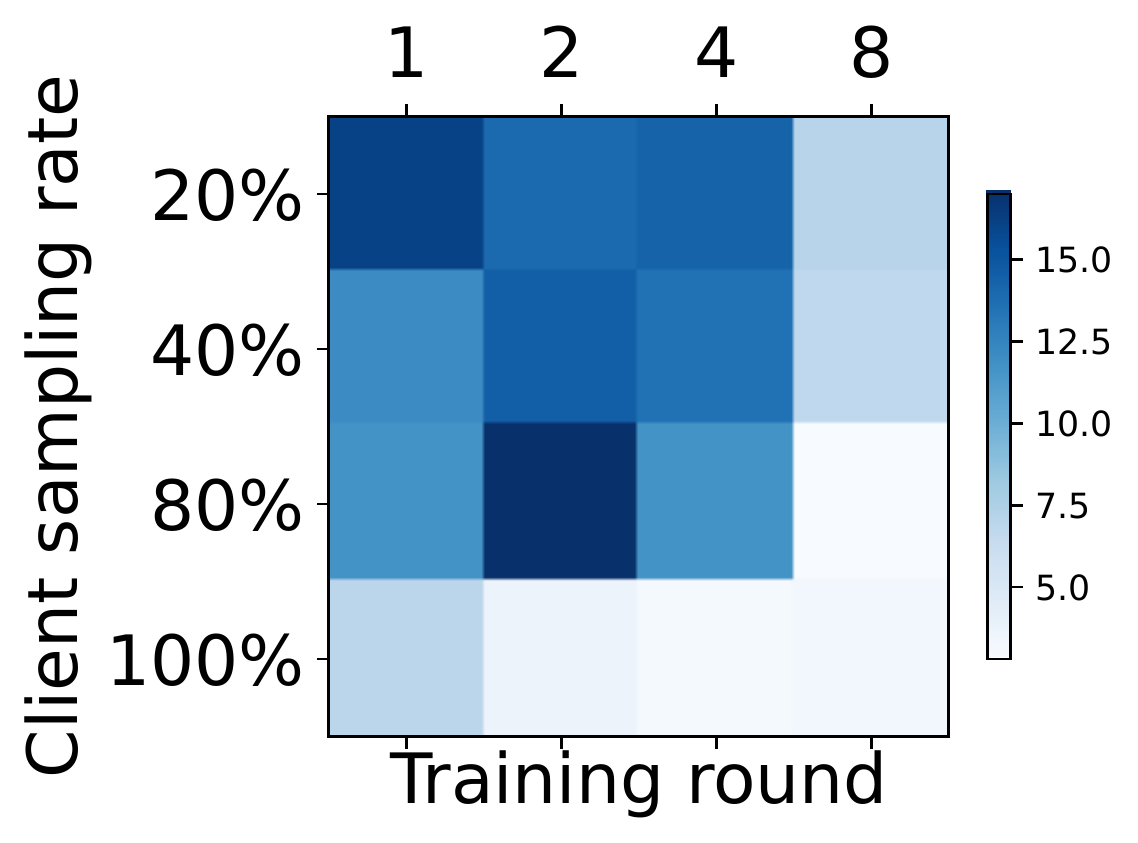}
    \caption{GCN.}
    \label{subfig:hpo_pubmed_gcn}
    \end{subfigure}
    \hfill
    \begin{subfigure}[b]{0.22\textwidth}
    \centering
    \includegraphics[width=\textwidth]{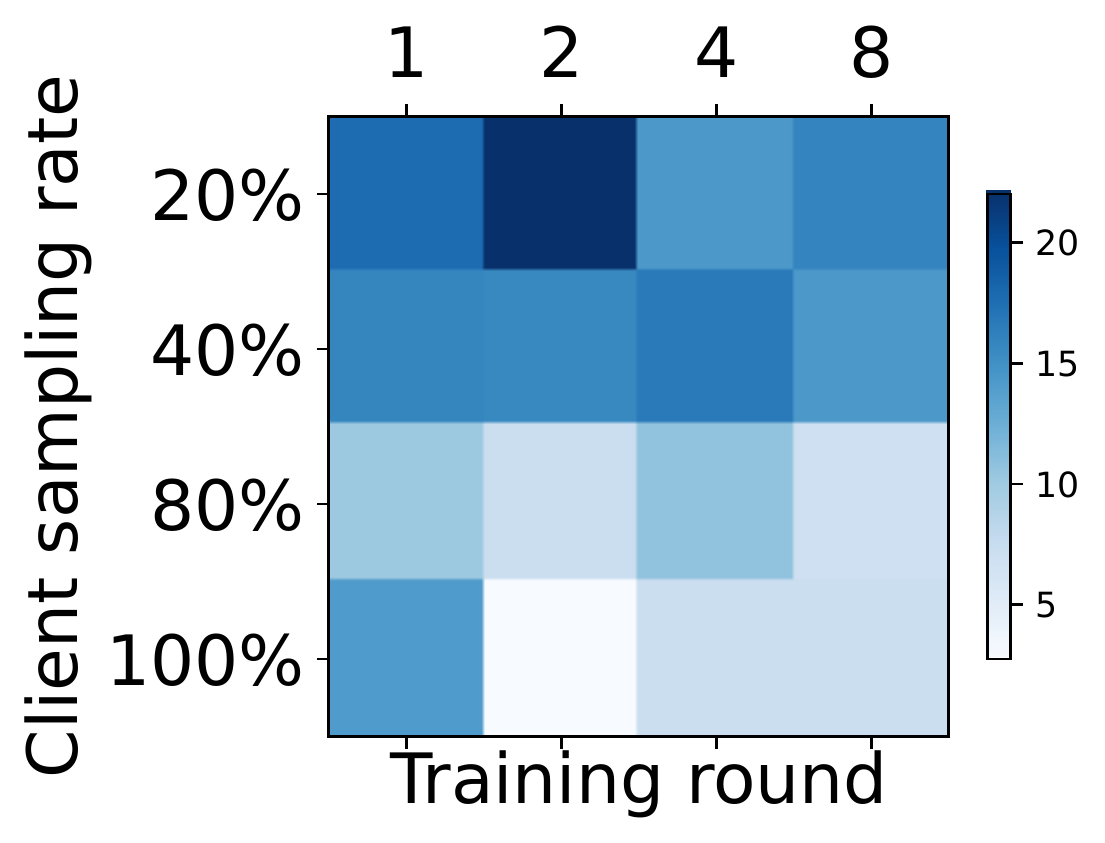}
    \caption{GPR-GNN.}
    \label{subfig:hpo_pubmed_gpr}
    \end{subfigure}
    \vspace{-0.1in}
    \caption{SHA with various fidelity to optimize GCN/GPR-GNN: We report the average ranking of searched hyper-parameter configuration (the small, the better).}
    \label{fig:hpo_pubmed}
    \vspace{-0.1in}
\end{figure}

\noindent\textbf{Results and Analysis}. There are in total 72 possible configurations, with each of which we conduct the \subjabbr procedure and thus acquire the ground-truth performances. These results become a lookup table, making comparing HPO methods efficient. We present the experimental results in Fig.~\ref{fig:hpo_pubmed}, where higher fidelity leads to better configuration for both kinds of GNNs. At first, we want to remind our readers that the left-upper region in each grid table corresponds to extremely low-fidelity HPO. Although their performances are worse than those in the other regions, they have successfully eliminated a considerable fraction of poor configurations. Meanwhile, increasing fidelity through the two aspects, i.e., client sampling rate and the number of training rounds, reveal comparable efficiency in improving the quality of searched configurations. This property provides valuable flexibility for practitioners to keep a fixed fidelity while trading-off between these two aspects according to their system status (e.g., network latency and how the dragger behaves). All these observations suggest the application of low-fidelity HPO to \subjabbr, as well as the effectiveness of \ours's model-tuning component.

\subsection{Study about Non-I.I.D.ness and Personalization}
\label{subsec:psnexp}
We aim to study the unique covariate shift of graph data---node attributes of different clients obey the identical distributions, but their graph structures are non-identical.
Meantime, we evaluate whether the personalization provided by \ours's model-tuning component can handle such non-i.i.d.ness.

\noindent\textbf{Protocol}. We choose \textit{FedcSBM} (see Sec.~\ref{subsec:newdata}) as the dataset. Specifically, we randomly generate a sample from our \textit{FedcSBM}, containing eight graphs with different homophilic degrees. Then we choose GPR-GNN as the GNN model and consider three settings: \textit{(1) Ordinary FL}: We apply FedAvg to optimize the model where the clients collaboratively optimize all model parameters; (2) \textit{Local}: We allow each client to optimize its model on its graph; (3) \textit{Personalization}: We apply FedAvg to optimize the parameters for feature transformation while the spectral coefficients (i.e., that for propagation) are client-specific. We repeat such a procedure five times and report the mean test accuracy and the standard deviation. More details can be found in Appendix~\ref{subsec:expdetail}.

\vspace{-0.1in}
\begin{figure}[th]
    \centering
    \includegraphics[width=0.375\textwidth]{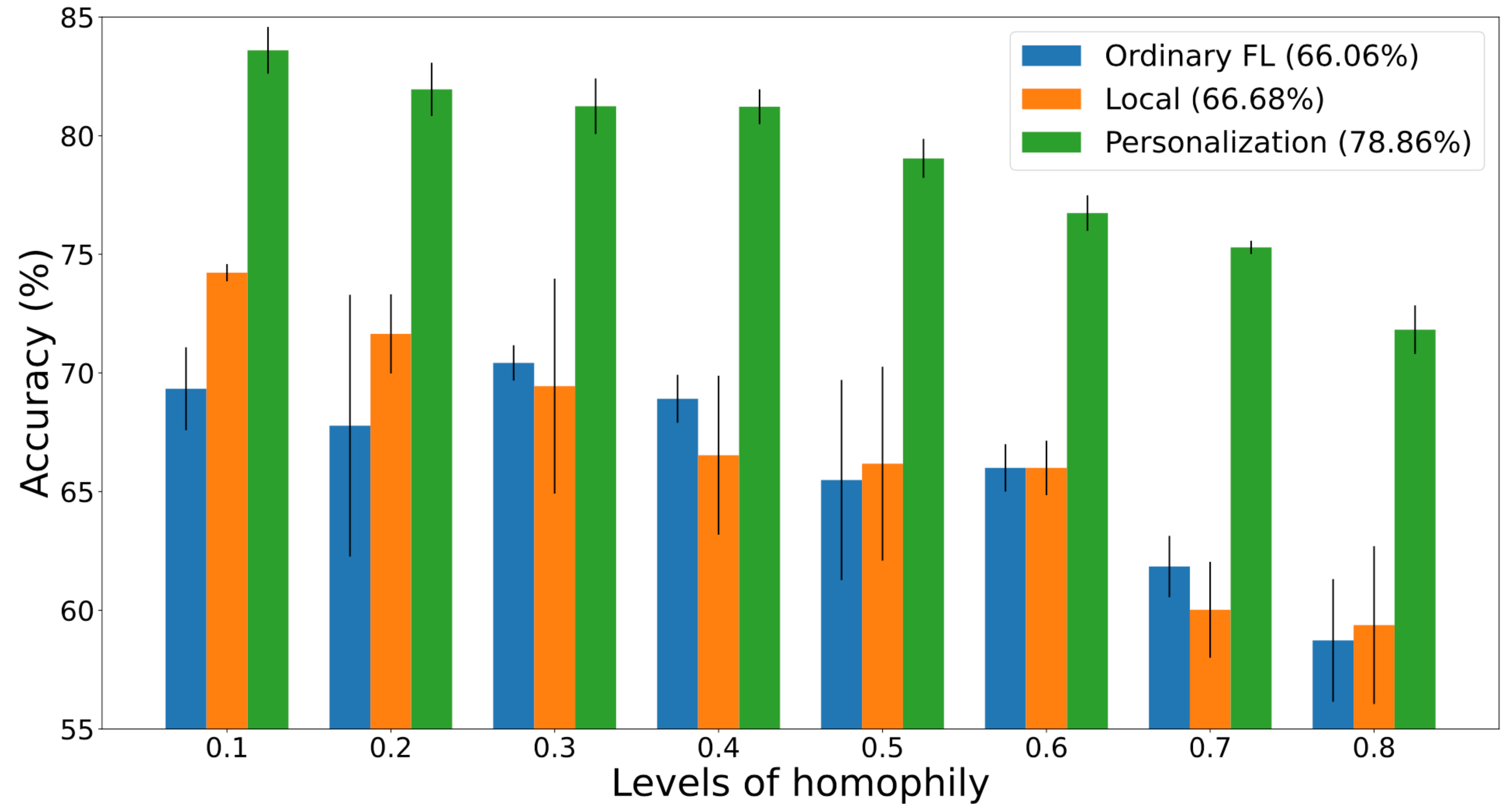}
    \vspace{-0.1in}
    \caption{Accuracy by levels of homophily and methods.}
    \vspace{-0.1in}
    \label{fig:accvshomo}
\end{figure}

\noindent\textbf{Results and Analysis}. We illustrate the results in Fig.~\ref{fig:accvshomo}, where each level of homophilic degree corresponds to a client. Overall, the personalization setting outperforms others. We attribute this advantage to making appropriate personalization, as the personalization setting consistently performs better across different clients, i.e., on graphs with different levels of homophilic degrees. On the other hand, the ordinary FL exhibits comparable performances with the local setting, implying the collaboration among clients fails to introduce any advantage. FedAvg might fail to converge due to the dissimilarity among received parameters, as the B-local dissimilarities shown in Fig.~\ref{subfig:dissim} indicate.

\vspace{-0.05in}
\subsection{Deployment in Real-world E-commerce Scenarios}
\label{subsec:realworld}
We deploy \ours to serve one federation with three E-commerce scenarios operated by different departments. These scenarios include search engine and recommender systems before and after the purchase behavior. As they all aim to predict whether each given user will click a specific item, and their users and items have massive intersections, sharing the user-item interaction logs to train a model is promising and has long been their solution. However, such data sharing raises the risk of privacy leakage and breaks the newly established regulations, e.g., Cybersecurity Law of China.

Without data sharing, each party has to train its model on its user-item bipartite graph, missing the opportunity to borrow strength from other graphs. Considering the scales of their graphs are pretty different, e.g., one scenario has only \num{184419} interactions while another has \num{2860074}, this might severely hurt the performance on "small" scenarios. With our \subjabbr service, these parties can collaboratively train a GraphSAGE model for predicting the link between each user-item pair. Evaluation on their held-out logs shows that the ROC-AUC of individually trained GraphSAGE models is 0.6209±0.0024 while that of FGL is 0.6278±0.0040, which is a significant improvement. It is worth noting that a small improvement (at 0.001-level) in offline ROC-AUC evaluation means a substantial difference in real-world click-through rate prediction for search/recommendation/advertisements. Therefore, the improvement confirms the business value of \ours.
\section{Conclusion}
\label{sec:conclusion}
In this paper, we implemented an \subjabbr package, \ours, to facilitate both the research and application of \subjabbr. Utilizing \ours, \subjabbr algorithms can be expressed in a unified manner, validated against comprehensive and unified benchmarks, and further tuned efficiently. Meanwhile, \ours provides rich plug-in attack and defence utilities to assess the level of privacy leakage for the \subjabbr algorithm of interest. Besides extensive studies on benchmarks, we deploy \ours in real-world E-commerce scenarios and gain business benefits. We will release \ours to create greater business value from the ubiquitous graph data while preserving privacy.

\bibliographystyle{ACM-Reference-Format}
\bibliography{ref}

\clearpage
\appendix
\section{Appendix}
\label{sec:app}

\subsection{Details of Off-the-shelf Splitters}
\label{subsec:splitterdetails}
To this end, we have implemented mainly six classes of \textit{splitters}:

\textit{(1) community\_splitter}: This is often adopted in node-level tasks to simulate the locality-based federated graph data~\cite{fedsage}, where nodes in the same client are densely connected while cross-client edges are unavailable. Specifically, community detection algorithms (e.g., Louvain~\cite{louvain} and METIS~\cite{metis}) are at first applied to partition a graph into several clusters. Then these clusters are assigned to the clients, optionally with the objective of balancing the number of nodes in each client.

\textit{(2) random\_splitter}: Random split is often adopted in node-level tasks, e.g., FedGL~\cite{fedgl}. Specifically, the node set of the original graph is randomly split into $N$ subsets with or without intersections.
Then, the subgraph of each client is deduced from the nodes assigned to that client.
Optionally, a specified fraction of edges is randomly selected to be removed.

\textit{(3) meta\_splitter}: In many cases, there are meta data or at least interpretable edge/node attributes that allow users to simulate a real FL setting via splitting the graph based on the meta data or the values of those attributes. In citation networks, papers published in different conferences or organizations usually focus on different themes. Splitting by conference/organization naturally leads to node (i.e., paper) classification tasks with non-identical label distributions (i.e., prior probability shift~\cite{survey1}). Meanwhile, in recommender systems, the same user often has different tendencies to items in different domains/scenarios, where splitting by domain/scenario can provide concept shift among clients.

\textit{(4) instance\_space\_splitter}: It is responsible for creating feature distribution skew (i.e., covariate shift). To realize this, we sort the graphs based on their values of a certain aspect, e.g., for Molhiv, molecules are sorted by their scaffold, and then each client is assigned with a segment of the sorted list.

\textit{(5) label\_space\_splitter}: It is designed to provide label distribution skew. For classification tasks, e.g., relation prediction for knowledge graph completion, the existing triplets are split into the clients by latent dirichlet allocation (LDA)~\cite{LDA}. For regression tasks, e.g., PCQM4M, \ours can discretize the label before conducting LDA.

\textit{(6) multi\_task\_splitter}: This is mainly designed for multi-task learning or personalized learning. Sometimes different clients have different tasks, e.g., in the domain of the molecule, some clients have the task of determining the toxicity, while some clients have the task of predicting the HOMO-LUMO gap. A more challenging case~\cite{gcfl} is that the graphs come from the different domains, e.g., molecules, proteins, and social networks.

\begin{table*}[tb]
\centering
\caption{Datasets statistics.}
\label{tab:datasets}
\resizebox{\textwidth}{!}{
\begin{tabular}{ccccccccc}
\hline
Task                         & Domain                & Dataset   & Splitter            & \# Graph & Avg. \# Nodes & Avg. \# Edges & \# Class & Evaluation \\ \hline
\multirow{4}{*}{Node-level} & Citation network      & Cora & \textit{random\&community} & 1  & \num{2708}    & \num{5429}     & 7 & ACC \\
                             & Citation network      & CiteSeer  & \textit{random\&community} & 1        & \num{4230}          & \num{5358}          & 6        & ACC       \\
                             & Citation network      & PubMed    & \textit{random\&community} & 1        & \num{19717}         & \num{44338}         & 5        & ACC       \\
                             & Citation network      & FedDBLP   & \textit{meta}                & 1        & \num{52202}         & \num{271054}        & 4        & ACC       \\ \hline
\multirow{4}{*}{Link-level} & Recommendation System & Ciao & \textit{meta}                & \num{28} & \num{5875.68} & \num{20189.29} & 6 & ACC \\
                             & Recommendation System & Taobao    & \textit{meta}                & 3        & \num{443365}        & \num{2015558}       & 2        & ACC       \\
                             & Knowledge Graph       & WN18      & \textit{label\_space}        & 1        & \num{40943}         & \num{151442}        & \num{18}       & Hits@n      \\
                             & Knowledge Graph       & FB15k-237 & \textit{label\_space}        & 1        & \num{14541}         & \num{310116}        & \num{237}      & Hits@n      \\ \hline
\multirow{4}{*}{Graph-level} & Molecule              & HIV       & \textit{instance\_space}     & \num{41127}    & 25.51         & 54.93         & 2        & ROC-AUC    \\
                             & Proteins              & Proteins  & \textit{instance\_space}     & \num{1113}     & \num{39.05}         & \num{145.63}        & 2        & ACC       \\
                             & Social network        & IMDB      & \textit{label\_space}        & \num{1000}     & \num{19.77}         & \num{193.06}        & 2        & ACC       \\
                             & Multi-task            & Mol       & \textit{multi\_task}         & \num{18661}    & \num{55.62}         & \num{1466.83}       & -        & ACC       \\ \hline
\end{tabular}
}
\end{table*}

\subsection{Details about Our Experiments}
\label{subsec:expdetail}
\noindent\textbf{Experimental settings}. In node-level tasks, the detailed hyper-parameters in our experiments are as follows: the number of training rounds is 400, the early stopping is 100, the GNN layers is 2 (in GPR-GNN, K is 10), the hidden layer dimension is 64 on citation networks and 1024 on FedDBLP, the dropout is 0.5, weight decay is 0.0005,  the number of clients is five on citation networks and the optimizer is SGD.

\noindent\textbf{More results and analysis about node-level tasks}. Particularly, we consider one of the recently proposed \subjabbr algorithms, FedSAGE+~\cite{fedsage}, which is highlighted by simultaneously training generative models for predicting the missing links so that its GraphSAGE models can be trained on the mended graphs. We show the results in Table~\ref{tab:fedsage}, where FedSage+ significantly outperforms its baseline (i.e., GraphSage) on most of the datasets. It benefits from the jointly learned generative models, which enable each client to reconstruct the missing cross-client links under federated setting.

We present the results on \textit{FedDBLP} in Table~\ref{tab:nodelevel_meta}, where the fraction of removed edges reaches 60\% and 40\% when split by venue and organizer, respectively. Besides, there are 20 and 8 clients under the two splitting settings, respectively, larger than the previous experiments. Since the client-specific graphs are tiny, w.r.t. the original one, the available training examples are limited for the local setting. All these factors make the performances of different GNNs unsatisfactory under the local setting. As for \subjabbr, since FedAvg aggregates the clients' updates, it somehow exploits all the training examples and thus achieves comparable performances against the global setting. Considering that \textit{FedDBLP} has simulated the data interruption in real life, these results confirm the effectiveness of \subjabbr to handle this emerging challenge.

\noindent\textbf{More results and analysis about link-level tasks}. We provide more experimental results on the FedDBLP, Ciao, and HIV in the Table.~\ref{tab:nodelevel_meta}, Table.~\ref{tab:linklevel_rec}, and Table.~\ref{tab:graphlevel_hiv}. All experimental settings are consistent with Sec.~\ref{subsubsec:node-level-tasks}, Sec.~\ref{subsubsec:link-level-tasks}, and Sec.~\ref{subsubsec:graph-level-tasks}, respectively.

\noindent\textbf{More results and analysis about graph-level tasks}. In addition, we consider the recently proposed \subjabbr algorithm GCFL+~\cite{gcfl}, which clusters clients and performs FedAvg in a cluster-wise manner so that clients with similar data distributions share a common GIN model. In Table~\ref{tab:gcfl}, GCFL+ outperforms its baseline (i.e., GIN federally learned by FedAvg) on both the IMDB and Multi-task datasets, and achieves comparable performance on PROTEINS. GCFL+ clusters the clients according to their sequences of gradients, where clients belonging to the same cluster share the same model parameters. Its advantages are likely to come from this smart mechanism, which handles the non-i.i.d.ness among clients better.

\begin{table}[htbp]
\centering
\caption{Comparisons between FedSage+ and GraphSAGE (with FedAvg) on representative node classification datasets: Mean accuracy (\%) ± standard deviation.}
\label{tab:fedsage}

\resizebox{0.475\textwidth}{!}{
\begin{tabular}{l|ll|ll|ll}
\hline
          & \multicolumn{2}{c|}{Cora}         & \multicolumn{2}{c|}{CiteSeer}     & \multicolumn{2}{c}{PubMed}        \\
          & \multicolumn{1}{c}{random}                & \multicolumn{1}{c|}{community} & \multicolumn{1}{c}{random}                & \multicolumn{1}{c|}{community} & \multicolumn{1}{c}{random}                & \multicolumn{1}{c}{community} \\ \hline
GraphSAGE & \multicolumn{1}{c|}{85.42±1.80} & 87.19±1.28           & \multicolumn{1}{c|}{76.86±1.38} & 77.80±1.03          & \multicolumn{1}{c|}{86.45±0.43} & 86.87±0.53          \\
FedSage+  & \multicolumn{1}{c|}{85.07±1.20} & 87.68±1.55          & \multicolumn{1}{c|}{78.04±0.91} & 77.98±1.23          & \multicolumn{1}{c|}{88.19±0.32 } & 87.94±0.27          \\ \hline
\end{tabular}
}
\end{table}

\begin{table}[htbp]
    \centering
    \caption{Results on representative node classification datasets with \textit{meta\_splitter}: Mean accuracy (\%) ± standard deviation.}
    \label{tab:nodelevel_meta}
    \resizebox{0.48\textwidth}{!}{
    \begin{tabular}{c|c|c|c|c|c|c}
    \hline
    & \multicolumn{3}{c|}{FedDBLP (by venue)} & \multicolumn{3}{c}{FedDBLP (by publisher)} \\
    & \multicolumn{1}{c}{Local} & \multicolumn{1}{c}{\subjabbr} & \multicolumn{1}{c|}{Global} & \multicolumn{1}{c}{Local} & \multicolumn{1}{c}{\subjabbr} & \multicolumn{1}{c}{Global} \\
    \hline
    GCN &  58.86±0.32 & 77.53±0.03 & 78.50±0.04 & 67.59±0.44 & 77.98±0.04 & 78.50±0.04 \\
    GraphSAGE & 51.05±0.43 & 78.57±0.03 & 78.96±0.39 &  61.60±0.19 & 79.23±0.05 & 78.96±0.39 \\
    GAT & 59.06±0.48 & 77.31±0.07 & OOM & 68.38±0.61 & 77.52±0.06 & OOM \\
    GPR-GNN & 58.07±0.39 &  76.53±0.23 & OOM & 66.10±0.25 &  78.17±0.12 & OOM \\
    \hline
    \end{tabular}
    }
\end{table}

\begin{table}[htbp]
    \centering
    \small
    \caption{Results on link classification dataset \textit{Ciao} with \textit{meta\_splitter}: Mean accuracy (\%) ± standard deviation.}
    \label{tab:linklevel_rec}
    \begin{tabular}{c|c|c|c}
    \hline
    & \multicolumn{3}{c}{Ciao} \\
    & \multicolumn{1}{c}{Local} & \multicolumn{1}{c}{\subjabbr} & \multicolumn{1}{c}{Global} \\
    \hline
    GCN & 46.76±0.44 & 49.18±0.00 & 49.18±0.00 \\
    GraphSAGE & 46.62±0.35 & 49.18±0.00 & 49.18±0.00 \\
    GAT & 46.83±0.31 & 49.18±0.00 & 49.18±0.00 \\
    GPR-GNN & 47.73±0.75 & 49.24±0.08 & 49.21±0.07 \\
    \hline
    \end{tabular}
\end{table}

\begin{table}[htbp]
    \centering
    \small
    \caption{Results on graph classification dataset \textit{HIV} with \textit{instance\_space\_splitter}: Mean ROC-AUC scrore ± standard deviation.}
    \label{tab:graphlevel_hiv}
    \begin{tabular}{c|c|c|c}
    \hline
    & \multicolumn{3}{c}{HIV (ROC-AUC)} \\
    & \multicolumn{1}{c}{Local} & \multicolumn{1}{c}{\subjabbr} & \multicolumn{1}{c}{Global}\\
    \hline
    GCN & 0.6193±0.0319 & 0.6263±0.0332 & 0.6939±0.0165 \\
    GIN & 0.6925±0.0354 & 0.7774±0.0195 & 0.7958±0.0200 \\
    GAT & 0.6192±0.0101 & 0.6287±0.0197 & 0.7034±0.0201 \\
    \hline
    \end{tabular}
\end{table}

\begin{table}[htbp]
\small
\caption{Comparisons between GCFL+ and GIN (with FedAvg) on graph classification datasets: Mean accuracy (\%) ± standard deviation.}
\label{tab:gcfl}
\centering
\begin{tabular}{c|c|c|c}
\hline
      & \multicolumn{1}{c|}{PROTEINS} & \multicolumn{1}{c|}{IMDB} & \multicolumn{1}{c}{Multi-task} \\ \hline
GIN   & 73.74±5.71                               &  64.79±10.55                          &  63.40±2.22                              \\
GCFL+ &  73.00±5.72                              &  69.47±8.71                         & 65.14±1.23                               \\ \hline
\end{tabular}
\end{table}

\noindent\textbf{More details about the protocol of our studies on hyper-parameter optimization}. We largely follow the experimental setup in the Sec.~\ref{subsubsec:node-level-tasks} while extend the search space of the hyper-parameters, where hidden dim $\in\{32, 64\}$, dropout $\in\{0.0, 0.5\}$, and weight decay in $\in\{0, 0.0005\}$. As PubMed has much more nodes and edges than the other two citation networks, we target the node classification task on PubMed to draw statistically reliable conclusions. In order to simulate the FL setting, we apply our \textit{community\_splitter} to divide the PubMed into five parts for five clients.

\noindent\textbf{Non-I.I.D.ness and Personalization Study}. In order to generate different level of homophily graphs, we set $\phi\in\{0.1, 0.2, 0.3, 0.4, 0.5,$ $0.6, 0.7, 0.8\}$ for cSBM model. We repeat our experiment with GPR-GNN for three-time with a different seed. In \subjabbr setting, each client shares the parameters of linear layers, while the parameters of the label propagation layer are personalized.

\noindent\textbf{Real-world Deployment}. We provide more statistical information about the real-world E-commerce scenarios dataset. Search engine scenario contains \num{106222} users and \num{464904} items. In the other two scenarios, the first contains \num{12588} users and \num{78996} items, and the second contains \num{107589} users and \num{559796}, respectively.

\subsection{Datasets description}
\label{subsec:Datasets Description}
As in Table~\ref{tab:datasets}, we provide a detailed description of datasets of \ours with datasets and suggested \textit{Splitter} accordingly. The datasets are collected from different domains, and the nodes and edges represent different meanings. We will support more datasets and provide more benchmarks in the future.

\end{document}